\documentclass[utf8]{frontiersSCNS} 

\usepackage{url,hyperref,lineno,microtype,subcaption}
\usepackage[onehalfspacing]{setspace}
\usepackage{multirow}
\linenumbers

\def\keyFont{\fontsize{8}{11}\helveticabold }
\def\firstAuthorLast{Chakraborty {et~al.}} 
\def\Authors{Biswadeep Chakraborty\,$^{1,*}$, Saibal Mukhopadhyay\,$^{1}$}
\def\Address{$^{1}$Georgia Institute of Technology, Department of Electrical and Computer Engineering, Atlanta , Georgia , USA }
\def\corrAuthor{Biswadeep Chakraborty}

\def\corrEmail{biswadeep@gatech.edu}

\newcommand{\hlt}[1]{ {\color{black}#1}}

\begin{document}
\onecolumn
\firstpage{1}

\title[Characterization of Generalizability of STDP]{Characterization of Generalizability of Spike \hlt{Timing} Dependent Plasticity trained Spiking Neural Networks} 

\author[\firstAuthorLast ]{\Authors} 
\address{} 
\correspondance{} 

\extraAuth{}

\maketitle

\begin{abstract}

A Spiking Neural Network (SNN) \hlt{is trained with Spike Timing} Dependent Plasticity (STDP), which is a neuro-inspired unsupervised learning method for various machine learning applications. This paper studies the generalizability properties of the STDP learning processes using the Hausdorff dimension of the trajectories of the learning algorithm.
The paper analyzes the effects of STDP learning models and associated hyper-parameters on the generalizability properties of an SNN.
The analysis is used to develop a Bayesian optimization approach to optimize the hyper-parameters for an STDP model \hlt{for improving} the generalizability properties of an SNN.

\tiny
 \keyFont{ \section{Keywords:}Spiking Neural Networks, Leaky Integrate and Fire, Generalization, Hausdorff Dimension, logSTDP, addSTDP, multSTDP, Bayesian Optimization} 
\end{abstract}

\section{Introduction}

A Spiking Neural Network (SNN) ~\citep{maass1997networks, gerstner2002spiking, pfeiffer2018deep} is a neuro-inspired machine learning (ML) model that mimics the spike-based operation of the human brain ~\citep{bi1998synaptic}. 
The Spike \hlt{Timing} Dependent Plasticity (STDP) is a policy for unsupervised learning in SNNs~\citep{bell1997synaptic, magee1997synaptically, gerstner2002mathematical}. The STDP relates the expected change in synaptic weights to the timing difference between \hlt{post-synaptic} spikes and \hlt{pre-synaptic} spikes ~\citep{feldman2012spike}. Recent works using STDP trained SNNs have demonstrated promising results as an unsupervised learning paradigm for various tasks such as object classification and recognition~\citep{she2021heterogeneous, diehl2015unsupervised, kheradpisheh2018stdp, masquelier2009competitive, lee2018deep, mozafari2019bio}.

The generalizability is a measure of how well an ML model performs on test data that lies outside of the distribution of the training samples ~\citep{kawaguchi2017generalization, neyshabur2017exploring}. The generalization properties of  Stochastic Gradient Descent (SGD) based training for deep neural network (DNN) have received significant attention in recent years ~\citep{poggio2019theoretical, allen2018learning, allen2019can}. 
The dynamics of SGD have been studied via models of stochastic gradient Langevin dynamics with an assumption that gradient noise is Gaussian ~\citep{simsekli2020fractional}. Here SGD is considered to be driven by a Brownian motion.  Chen et al. showed that SGD dynamics commonly exhibit exhibits rich, complex dynamics when
navigating through the loss landscape ~\citep{chen2020anomalous}. \hlt{Recently Gürbüzbalaban et al. ~\citep{gurbuzbalaban2020heavy} and Hodgkinson et al. \citep{hodgkinson2021multiplicative} have simultaneously shown that the law of the SGD iterates can converge to a heavy-tailed stationary distribution with
infinite variance when the step-size $\eta$ is large and/or the batch-size $B$ is small. These results form a theoretical
basis for the origins of the observed heavy-tailed behavior of SGD in practice.} 
The authors proved generalization bounds for SGD under the assumption that its trajectories can be well-approximated by the \hlt{Feller Process \citep{capocelli1976transformation}}, a Markov-based stochastic process. Modeling the trajectories of SGD using a stochastic differential equation (SDE) under heavy-tailed gradient noise has shed light on several interesting characteristics of SGD.

In contrast, the generalizability analysis of STDP trained SNNs, although important, has received much less attention. Few studies have shown that, in general, the biological learning process in the human brain has significantly good generalization properties ~\citep{zador2019critique,sinz2019engineering}. However, none of them have characterized the generalization of an STDP-trained SNN using a mathematical model. There is little understanding of how hyperparameters of the STDP process impact the generalizability of the trained SNN model. Moreover, the generalization of STDP cannot be characterized by directly adopting similar studies for SGD. For example, SGD has been modeled as a Feller process for studying generalizability.

\hlt{Rossum et al. showed that random variations arise due to the variability in the amount of potentiation (depression) between the pre-and post-synaptic events at fixed relative timing  \citep{van2000stable}. At the neuron level, fluctuations in relative timing between pre-and post-synaptic events also contribute to random variations \citep{roberts2000computational}. 
 For many STDP learning rules reported in the literature, the dynamics instantiate a Markov process (\citep{bell1997synaptic}; \citep{markram1997regulation}; \citep{bi1998synaptic}; \citep{han2000reversible}; \citep{van2000stable}); changes in the synaptic weight depend through the learning rule only on the current weight and a set of random variables that determine the transition probabilities. However, recent literature have shown that weight update using STDP is better modeled as an Ornstein-Uhlenbeck process ~\citep{cateau2003stochastic}, ~\citep{aceituno2020spiking}, ~\citep{legenstein2008learning}. }

\hlt{
As described by Camuto et al. \citep{camuto2021fractal}, fractals are complex patterns, and the level of this complexity is typically measured by the Hausdorff dimension (HD)
of the fractal, which is a notion of dimension. Recently, assuming that SGD trajectories can be well-approximated by a Feller Process, it is shown that the generalization error can be controlled by the Hausdorff dimension of the trajectories of the SDE \cite{simsekli2020hausdorff}. That is, the
ambient dimension that appears in classical learning theory bounds is replaced with the Hausdorff dimension.
The fractal geometric approach presented by Simsekli et al. can capture the low dimensional structure of fractal sets and provides an alternative perspective to the compression-based approaches that aim to understand why over parametrized networks do not overfit.

}

This paper presents a model to characterize the generalizability of the STDP process and develops a methodology to optimize hyperparameters to improve the generalizability of an STDP-trained SNN. We use the fact that the sample paths of a Markov process exhibit a fractal-like structure ~\citep{xiao2003random}. The generalization error over the sample paths is related to the roughness of the random fractal generated by the driving Markov process which is measured by the Hausdorff dimension ~\citep{simsekli2020hausdorff} which is in turn dependent on the tail behavior of the driving process. 

\hlt{ 
Normally, validation loss of a model on a testing set is used to characterize the accuracy of that model. However, the validation loss is dependent on the choice of the test set, and does not necessarily give a good measure of the generalization of the learning process. Therefore, generalization error a model is generally measured normally measured by comparing the difference between training and testing accuracy - a more generizable model has less difference between training and testing accuracy ~\citep{goodfellow2016deep}. However, such a measure of generalization error requires computing the validation loss (i.e. testing accuracy) for a given test dataset.  To optimize the generalizability of the model, we need an objective function that measures the generalizability of the learning process. If the validation loss is used as a measure, then for each iteration of the optimization, we need to compute this loss by running the model over the entire dataset, which will significantly increase the computation time. We use Hausdorff Dimension as a measure of generalizability of the STDP process to address the preceding challenges. 
First,  the Hausdorff measure characterizes the fractal nature of the sample paths of the learning process itself and does not depend on the testing dataset. Hence, HD provides a better (i.e. testset independent) measure of the generalization of the learning process. Second, HD can be computed during the training process itself and does not require running inference on the test data set. This reduces computation time per iteration of the optimization process.

}

We model the STDP learning as an Ornstein-Uhlenbeck process which is a Markov process and show that the generalization error is dependent on the Hausdorff dimension of the trajectories of the STDP process.
We use the SDE representation of synaptic plasticity and model STDP learning as a stochastic process that solves the SDE. 

Using the Hausdorff dimension we study the generalization properties of an STDP trained SNN for image classification. We compare three different STDP processes, namely, log-STDP, add-STDP, and mult-STDP, and show that the log-STDP improves generalizability. We show that modulating hyperparameters of the STDP learning rule and learning rate changes the generalizability of the trained model. Moreover, using log-STDP as an example, we show the hyperparameter choices that reduce generalization error increases the convergence time, and training loss, showing a trade-off between generalizability and the learning ability of a model. 
Motivated by the preceding observations, we develop a Bayesian optimization technique for determining the optimal set of hyperparameters which gives an STDP model with the least generalization error. \hlt{We consider an SNN model with 6400 learning neurons trained using the log-STDP process. Optimizing the hyperparameters of the learning process using Bayesian Optimization shows a testing accuracy of $90.65\%$  and a generalization error of $3.17$ on the MNIST dataset. This shows a mean increase of almost 40\% in generalization performance for a mean drop of about 1\% in testing accuracy in comparison to randomly initialized training hyperparameters.
In order to further evaluate the learning methodologies, we also evaluated them on the more complex Fashion MNIST dataset and observed a similar trend.
}

\section{MATERIALS AND METHODS}

\subsection{Background}

\subsubsection{Spiking Neural Networks}

We chose the leaky integrate-and-fire model of a neuron where the membrane voltage $X$ is described by
$$
\tau \frac{d X}{d t}=\left(E_{\text {rest }}-X\right)+g_{e}\left(E_{\text {exc }}-X\right)+g_{i}\left(E_{\text {inh }}-X\right)
$$
where $E_{\text {rest }}$ is the resting membrane potential; $E_{e x c}$ and $E_{\text {inh }}$ are the equilibrium potentials of excitatory and inhibitory synapses, respectively; and $g_{e}$ and $g_{i}$ are the conductances of excitatory and inhibitory synapses, respectively. The time constant $\tau,$ is longer for excitatory neurons than for inhibitory neurons. When the neuron's membrane potential crosses its membrane threshold, the neuron fires, and its membrane potential is reset. Hence, the neuron enters its refractory period and \hlt{cannot spike again for the duration of the refractory period.}

Synapses are modeled by conductance changes, i.e., synapses increase their conductance instantaneously by the synaptic weight $w$ when a \hlt{pre-synaptic} spike arrives at the synapse, otherwise, the conductance \hlt{decays} exponentially. Thus, the dynamics of the conductance $g$ can be written as:
\begin{equation}
    \tau_{g} \frac{d g}{d t}=-g
\end{equation}

\hlt{If the \hlt{pre-synaptic} neuron is excitatory, the dynamics of the conductance is $g = g_{e}$ with the time constant of the excitatory \hlt{post-synaptic} potential being $\tau_{g} = \tau_{g_{e}}$. On the other hand, if the \hlt{pre-synaptic} neuron is inhibitory, it's synaptic conductance is given as $g = g_{i}$ and the time constant of the inhibitory \hlt{post-synaptic} potential as $\tau_{g} = \tau_{g_{i}}$.}

\subsubsection{STDP based Learning Methods}
\label{sec:stdp}

Spike-timing-dependent plasticity is a biologically plausible learning model representing the time evolution of the synaptic weights as a function of the past spiking activity of adjacent neurons.

\begin{figure}
    \centering
    \includegraphics[width = \columnwidth]{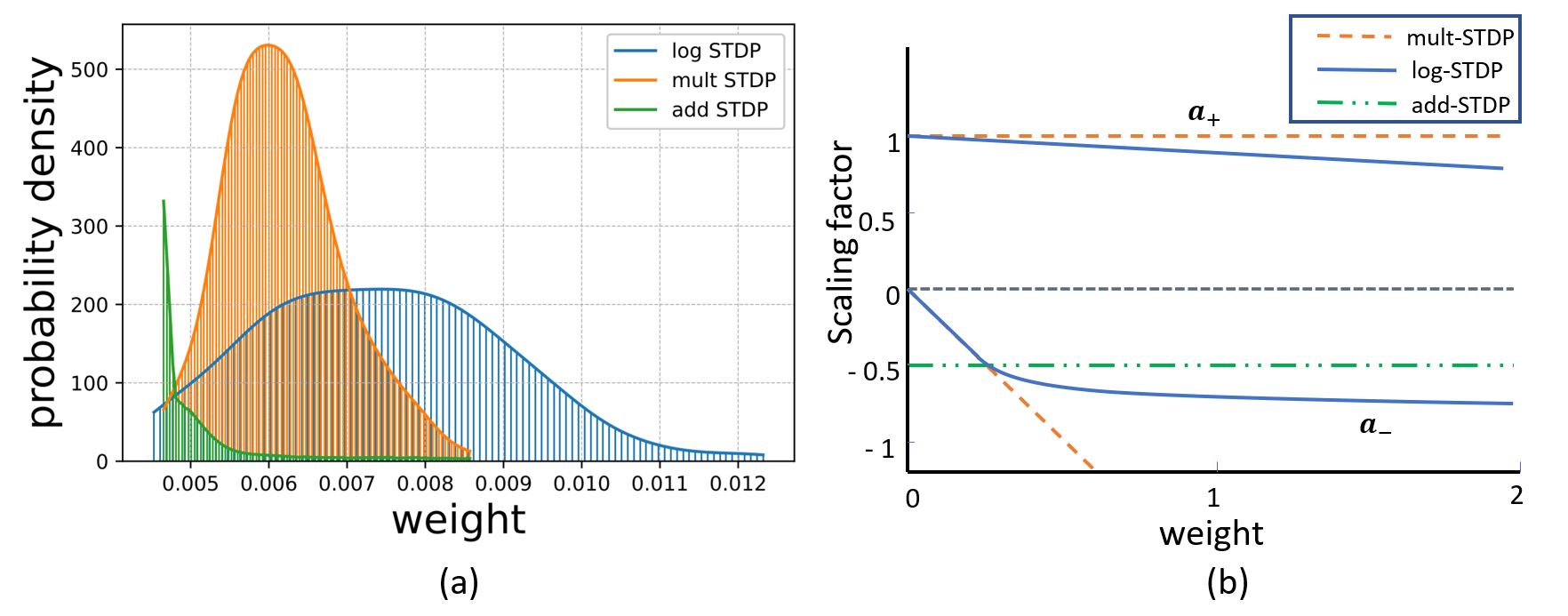}
    \caption{\hlt{(a) Resulting weight distribution for log-STDP ~\citep{gilson2011stability};multSTDP ~\citep{van2000stable} and add-STDP ~\citep{song2000competitive} (b) Plot for Functions $a_+$ for LTP and $-f_-$ for LTD in log-STDP (blue solid curve); mult-STDP (orange dashed line); and add-STDP model (green dashed-dotted curve for depression and orange dashed curve for potentiation}}
    \label{fig:stdp_comp}
\end{figure}

In a STDP model, the change in synaptic weight induced by the pre-and post-synaptic spikes at times $t_{pre}, t_{post}$ are defined by: 

\begin{equation}
\Delta W=\eta(1+\zeta) H\left(W ; t_{\text {pre }}-t_{\text {post }}\right)
\label{eq:synaptic_weight}
\end{equation}
where the learning rate $\eta$ determines the speed of learning. The Gaussian white noise $\zeta$ with zero mean and variance $\sigma^{2}$ describes the variability observed in physiology. The function $H(W ; u)$ describes the long term potentiation (LTP) and depression (LTD) based on the the relative timing of the spike pair within a learning window $u=t_{\text {pre }}-t_{\text {post }}$, and is defined by:

\begin{equation}
H(W ; u)=\left\{\begin{array}{ll}
a_{+}(W) \exp \left(-\frac{|u|}{\tau_{+}}\right) & \text {for } \mathrm{u}<0 \\
-a_{-}(W) \exp \left(-\frac{|u|}{\tau_{-}}\right) & \text {for } \mathrm{u}>0
\end{array}\right.
\label{eq:window}
\end{equation}
The shape of the weight distribution produced by STDP can be adjusted via the scaling functions $a_{\pm}(W)$ in \ref{eq:window} that determine the weight dependence. 
We study three different types of STDP processes, namely, add-STDP, mult-STDP, and log-STDP. All STDP models follow the equations \ref{eq:synaptic_weight} and \ref{eq:window}, however, they have different scaling functions ($a_\pm$) in \ref{eq:window} as discussed below. The weight distributions of these three different STDP processes at the end of the last training iteration are shown in Fig. 1.
At the beginning of the training iterations, the distribution is uniform for all three reflecting on the weight initialization conditions. Additive STDP gives rise to strong competition among synapses, but due to the absence of weight
dependence, it requires hard boundaries to secure the stability of weight dynamics. To reach a stability point for the add-STDP, we followed the analysis done by Gilson et al. \citep{gilson2011stability} and Burkitt et al. \citep{burkitt2004spike} and chose the fixed point $W_0 = 0.006$. Fig. 1 approximates the probability density function based on the weight distributions of the different STDP models. We are using a Gaussian KDE to get this pdf from the empirical weight distribution obtained.
Given $N$ independent realizations $\mathcal{X}_{N} \equiv \left\{X_{1}, \ldots, X_{N}\right\}$ from an unknown continuous probability density function (p.d.f.) $f$ on $\mathcal{X}$, the Gaussian kernel density estimator is defined as
\begin{equation}
    \hat{f}(x ; t)=\frac{1}{N} \sum_{i=1}^{N} \phi\left(x, X_{i} ; t\right), \quad x \in \mathbb{R}
\end{equation}

where
\begin{equation}
    \phi\left(x, X_{i} ; t\right)=\frac{1}{\sqrt{2 \pi t}} e^{-\left(x-X_{i}\right)^{2} /(2 t)}
\end{equation}

is a Gaussian p.d.f. (kernel) with location $X_{i}$ and scale $\sqrt{t}$. The scale is usually referred to as the bandwidth. Much research has been focused on the optimal choice of $t$, because the performance of $\hat{f}$ as an estimator of $f$ depends crucially on its value \citep{jones1992progress}, \citep{sheather1991reliable}.

\noindent \textbf{Logarithmic STDP (log-STDP)}~\citep{gilson2011stability}. 
The scaling functions of log-STDP is defined by:

\begin{align}
a_{+}(W)&=c_{+} \exp \left(-W / W_{0} \hlt{\gamma}\right) \\
a_{-}(W)&= \left\{\begin{array}{ll}c_{-} W / W_{0} \quad \text { for } \mathrm{W} \leq W_{0} \\
c_{-}\left[1+\frac{\ln \left[1+\hlt{\mathcal{S}}\left(\frac{W}{W_{0}}-1\right)\right]}{\hlt{\mathcal{S}}}\right] \text { for } \mathrm{W}>W_{0} \end{array}\right.
\label{eq:log_stdp}
\end{align}

 In this study, $W_{0}$ is chosen such that LTP and LTD in log-STDP balance each other for uncorrelated inputs, namely $A\left(W_{0}\right)=\tau_{+} a_{+}$ $\left(W_{0}\right)-\tau_{-} a_{-}\left(W_{0}\right) \simeq 0 .$ Therefore, $W_{0}$ will also be referred to as the 'fixed point' of the dynamics. 
 \hlt{Since depression increases sublinearly (blue solid curve for $a_{-}$ in Fig.\ref{fig:stdp_comp} ), noise in log-STDP is weaker than that for mult-STDP for which depression increases linearly (orange dashed curve for $a_{-}$ in Fig.\ref{fig:stdp_comp}).}

The weight dependence for LTD in logSTDP is similar to mult-STDP for $W \leq W_{0}$, i.e., it increases linearly with $W.$ However, the LTD curve $a_{-}$ becomes sublinear for $W \geq W_{0},$ and $\hlt{\mathcal{S}}$ determines the degree of the log-like saturation.  For larger $\hlt{\mathcal{S}}$, LTD has a more pronounced saturating log-like profile and the tail of the synaptic weight distribution extends further.

We choose the function $a_{+}$ for LTP to be roughly constant around $W_{0}$, such that the exponential decay controlled by $\hlt{\gamma}$ only shows for $W >> W_{0}$. 
\hlt{Thus, the scaling functions $a_+, a_-$, $\hlt{\mathcal{S}}$,and $\hlt{\gamma}$ are the hyperparameters that can be tuned to model an efficient log-STDP learning model.}

\noindent \textbf{Additive STDP (add-STDP)}~\hlt{\citep{song2000competitive}}. It is weight independent and is defined by the following scaling functions: 
    \begin{equation}
\begin{array}{l}
a_{+}(W)=c_{+} \\
a_{-}(W)=c_{-}
\end{array}
\end{equation}
with $c_{+} \tau_{+}<c_{-} \tau_{-}$ such that LTD overpowers LTP. The drift due to random spiking activity thus causes the weights to be depressed toward zero, which provides some stability for the output firing rate. \citep{gilson2011stability}.
\hlt{For the simulations, we are using a fast learning rate that is synonymous to a high level of noise, and more stability. This requires a stronger depression. Thus, we use $c_{+}=1$ and $c_{-}=0.6$.}

\noindent \textbf{Multiplicative STDP (mult-STDP)}~\citep{van2000stable}. The multiplicative STDP has a linear weight dependence for LTD and constant LTP:
\begin{align}
a_{+}(W)=c_{+} \\
a_{-}(W)=c_{-} W
\end{align}
The equilibrium mean weight is then given by $W_{\mathrm{av}}^{*}=c_{+} \tau_{+}$ $/\left(c_{-} \tau_{-}\right) .$ For the simulations we use $c_{+}=1$ and $c_{-}=0.5 / W_{0}=2$.  This calibration corresponds to a similar neuronal output firing rate to that for log-STDP in the case of uncorrelated inputs.

\subsubsection{Generalization - Hausdorff Dimension and Tail Index Analysis}

Recent works have discussed the generalizability of SGD based on the trajectories of the learning algorithm.  Simsekli et al. ~\citep{simsekli2020hausdorff} identified the complexity of the fractals generated by a Feller process that approximates SGD. The intrinsic complexity of a fractal is typically characterized by a notion called the Hausdorff dimension ~\citep{le2019hausdorff, lHorinczi2019multifractal}, which extends the usual notion of dimension to fractional orders. Due to their recursive nature, Markov processes often generate random fractals ~\citep{xiao2003random}. Significant research has been performed in modern probability theory to study the structure of such fractals  ~\citep{bishop2017fractals, khoshnevisan2009fractals, khoshnevisan2017macroscopic, yang2018multifractality}. 
\hlt{Thus, the STDP learning method follows an Ornstein-Uhlenbeck (O-U) process which is a special type of Lévy process. Again, the Hausdorff Dimension measures the roughness of the fractal patterns of the sample paths generated by the stochastic process which is measured using the tail properties of the  Lévy measure of the O-U process. 
 Lévy measure is a Borel measure on $\mathbb{R}^{d}\backslash \{0\}$ satisfying $\int_{\mathbb{R}^{d}}\|x\|^{2} /\left(1+\|x\|^{2}\right) \nu(\mathrm{d} x)<\infty$. The  Ornstein-Uhlenbeck process which is a  Lévy process can thus be characterized by the triplet $(b, \Sigma, \nu)$ where $b \in \mathbb{R}^d$ denotes a constant drift, $\Sigma \in \mathbb{R}^{d \times d}$ is a positive semi-definite matrix and $\nu$ is the Lévy measure as defined above. Thus, taking  Lévy processes as stochastic objects, their sample  path  behavior can be characterized by the Hausdorff dimension which is in turn measured using the BG-indices. Thus, the generalization properties of the STDP process can be modeled using the Hausdorff dimension of the sample paths of the O-U process. We formally define the Hausdorff dimension for the Ornstein-Uhlenbeck process modeling the STDP learning process in Section \ref{sec:hausdorff}.

}

\subsection{STDP as a Stochastic Process}
\label{sub:our-approach}

In this paper, we evaluate the generalization properties of an STDP model using the concept of the Hausdorff dimension. In this section, we discuss the learning methodology of STDP and how the plasticity change can be modeled using a stochastic differential equation. 
The state of a neuron is usually represented by its membrane potential $X$ which is a key parameter to describe the cell activity. Due to external input signals, the membrane potential of a neuron may rise until it reaches some threshold after which a spike is emitted and transferred to the synapses of neighboring cells. To take into account the important fluctuations within cells, due to the spiking activity and thermal noise, in particular, a random component in the cell dynamics has to be included in mathematical models describing the membrane potential evolution \hlt{of both the pre-and post-synaptic neurons} similar to the analysis shown by  Robert et al., ~\citep{robert2020stochastic}. Several models take into account this random component using an independent additive diffusion component, like Brownian motion, of the membrane potential $X$. In our model of synaptic plasticity, the stochasticity arises at the level of the generation of spikes. When the value of the membrane potential of the output neuron is at $X=x$, a spike occurs at rate $\beta(x)$ where $\beta$ is the activation function ~\citep{chichilnisky2001simple}. In particular, we consider the instants when the output neuron spikes are represented by an inhomogeneous Poisson process as used by Robert et al. ~\citep{robert2020stochastic}.
\hlt{Thus, in summary, (1) The pre-synaptic spikes are modeled using a Poisson process and hence, there is a random variable added to membrane potential. (2) The post-synaptic spikes are generated using a stochastic process based on the activation function. Hence, STDP, which depends on the pre-and post-synaptic spike times can be modeled using a  \textit{ stochastic differential equation (SDE)}.} Hence, we formulate the STDP as a SDE. The SDE of a learning algorithm shares similar convergence behavior of the algorithm and can be analyzed more easily than directly analyzing the algorithm.

\textbf{Mathematical Setup}: \hlt{We consider the STDP as an iterative learning algorithm $\mathcal{A}$ which is dependent on the dataset $\mathcal{D}$ and the algorithmic stochasticity $\hlt{\mathcal{U}}$. The learning process $\mathcal{A}(\mathcal{D},\hlt{\mathcal{U}})$ returns the entire evolution of the parameters of the network in the time frame $[0, T]$ where \hlt{$[\mathcal{A}(\mathcal{D},\hlt{\mathcal{U}})]_t = W_t$} being the parameter value returned by the STDP learning algorithm at time $t$. So, for a given training set $\mathcal{D}$, the learning algorithm $\mathcal{A}$ will output a random process ${w}_{t \in [0,T]}$ indexed by time which is a trajectory of iterates.

In the remainder of the paper, we will consider the case where the STDP learning process $\mathcal{A}$ is chosen to be the trajectories produced by the Ornstein-Uhlenbeck (O-U) process $\mathrm{W}^{(\mathcal{D})}$, whose symbol depends on the training set $\mathcal{D}$. More precisely, given $\mathcal{D} \in \mathcal{Z}^{n}$, the output of the training algorithm $\mathcal{A}(\mathcal{D}, \cdot)$ will be the random mapping $t \mapsto \mathrm{W}_{t}^{(\mathcal{D})}$, where the symbol of $\mathrm{W}^{(\mathcal{D})}$ is determined by the drift $b_{\mathcal{D}}(w)$, diffusion matrix $\Sigma_{\mathcal{D}}(w)$, and the Lévy measure $\nu_{\mathcal{D}}(w, \cdot)$, which all depend on $\mathcal{U}$. In this context, the random variable $\mathcal{U}$ represents the randomness that is incurred by the O-U process. Finally, we also define the collection of the parameters given in a trajectory, as the image of $\mathcal{A}(\mathcal{D})$, i.e., $\mathcal{W}_{\mathcal{D}}:=\left\{w \in \mathbb{R}^{d}: \exists t \in[0,1], w=[\mathcal{A}(\mathcal{D})]_{t}\right\}$ and the collection of all possible parameters as the
union $\mathcal{W}:=\bigcup_{n \geq 1} \bigcup_{\mathcal{D} \in \mathcal{Z}^{n}} \mathcal{W}_{S}$. Note that $\mathcal{W}$ is still random due to its dependence on $\mathcal{U}$.

We consider the dynamics of \hlt{synaptic} plasticity as a function of the membrane potential $X(t)$ and the synaptic weight $W(t)$. 

}

\textbf{Time Evolution of Synaptic Weights and Plasticity Kernels}
As described by Robert et al., ~\citep{robert2020stochastic}, the time evolution of the weight distribution $W(t)$ depends on the past activity of the input and output neurons. It may be represented using the following differential equation:
\begin{equation}
    \frac{dW(t)}{dt} = M(\Omega_{p}(t) , \Omega_{d}(t), W(t) )
    \label{eq:sde1}
\end{equation}
where $\Omega_{p}(t), \Omega_{d}(t)$ are two non-negative processes where the first one is associated with potentiation i.e., increase in W and the latter is related to the depression i.e., decrease in W. 
 \hlt{The function $M$ needs to be chosen such that the synaptic weight $W$ stays at all-time in its definition interval $K_W$. The function $M$ can thus be modified depending on the type of implementation of STDP that is needed. Further details regarding the choice of $M$ for different types of STDP is discussed by Robert et al.~\citep{robert2020stochastic} 
 
 }
 When the synaptic weight of a connection between a pre-synaptic neuron and a post-synaptic neuron is fixed and equal to $W$, the time evolution of the post-synaptic membrane potential $X(t)$ is represented by the following stochastic differential equation (SDE) ~\citep{robert2020stochastic}:

\begin{equation}
\mathrm{d} X(t)=-\frac{1}{\tau} X(t) \mathrm{d} t+W \mathcal{N}_{\lambda}(\mathrm{d} t)-g(X(t-)) \mathcal{N}_{\beta, X}(\mathrm{~d} t)
\label{eq:sde2}
\end{equation}

where $X(t-)$ is the left limit of $X$ at $t>0$, and $\tau$ is the exponential decay time constant of the membrane potential associated with the leaking mechanism. The sequence of firing instants of the pre-synaptic neuron is assumed to be a Poisson point process $\mathcal{N}_{\lambda}$ on $\mathbb{R}_{+}$ with the rate $\lambda$. At each pre-synaptic spike, the membrane potential $X$ is increased by the amount $W .$ If $W>0$ the synapse is said to be excitatory, whereas for $W<0$ the synapse is inhibitory. The sequence of firing instants of the post-synaptic neuron is an inhomogeneous Poisson point process $\mathcal{N}_{\beta, X}$ on $\mathbb{R}_{+}$ whose intensity function is $t \mapsto \beta(X(t-))$. The drop of potential due to a post-synaptic spike is represented by the function $g.$ When the post-synaptic neuron fires in-state $X(t-)=x$, its state $X(t)$ just after the spike is $x-g(x)$.

\hlt{\textbf{Uniform Hausdorff Dimension:} The Hausdorff dimension for the training algorithm $\mathcal{A}$ is a notion of complexity based on the trajectories generated by $\mathcal{A}$. Recent literature has shown that the synaptic weight update using an STDP rule can be approximated using a type of stochastic process called the Ornstein-Uhlenbeck process which is a type of Markov process ~\citep{cateau2003stochastic}, ~\citep{aceituno2020spiking}, ~\citep{legenstein2008learning}. Hence, we can infer that the STDP process will also have a uniform Hausdorff dimension for the trajectories. We use the Hausdorff Dimension of the sample paths of the STDP based learning algorithm which has not been investigated in the literature.

 Let $\Phi$ be the class of functions $\varphi:(0, \delta) \rightarrow(0, \infty)$ which are right continuous, monotone increasing with $\varphi(0+)=0$ and such that there exists a finite constant $K>0$ such that
\begin{equation}
    \frac{\varphi(2 s)}{\varphi(s)} \leq K, \quad \text { for } 0<s<\frac{\delta}{2} 
\label{eq:haus0}
\end{equation}

A function $\varphi$ in $\Phi$ is often called a measure function. For $\varphi \in \Phi$, the $\varphi$-Hausdorff measure of $E \subseteq \mathbb{R}^{d}$ is defined by
\begin{equation}
    \varphi_m(E)=\lim _{\varepsilon \rightarrow 0} \inf \left\{\sum_{i} \varphi\left(2 r_{i}\right): E \subseteq \bigcup_{i=1}^{\infty} B\left(x_{i}, r_{i}\right), r_{i}<\varepsilon\right\}
    \label{eq:haus1}
\end{equation}

where $B(x, r)$ denotes the open ball of radius $r$ centered at $x .$ The sequence of balls satisfying Eq. \ref{eq:haus1} is called an $\varepsilon$ -covering of $E$. We know that $\varphi_m$ is a metric outer measure and every Borel set in $\mathbb{R}^{d}$ is $\varphi_m$ measurable. Thus, the function $\varphi \in \Phi$ is called the Hausdorff measure function for $E$ if $0<\varphi_m(E)<\infty$.

It is to be noted here that in Eq. \ref{eq:haus1}, we only use coverings of $E$ by balls, hence $\varphi_m$ is usually called a spherical Hausdorff measure in the literature. Under Eq. \ref{eq:haus0}, $\varphi_m$ is equivalent to the Hausdorff measure defined by using coverings by arbitrary sets.
The Hausdorff dimension of $E$ is defined by
\begin{equation}
    \operatorname{dim}_{\mathrm{H}} E=\inf \left\{\alpha>0: s^{\alpha}-m(E)=0\right\} .
\end{equation}

\hlt{
The STDP learning process is modeled using the SDEs for the temporal evolution of the synaptic weights and the membrane potential given in Eqs. \ref{eq:sde1}, \ref{eq:sde2}. Considering the empirical observation that STDP exhibits a diffusive behavior around a local minimum \citep{baity2018comparing}, we take $w_{\mathcal{D}}$ to be the local minimum found by STDP and assume that the conditions of Proposition hold around that point. This perspective indicates that the generalization error can be controlled by the BG index $\beta_{\mathcal{D}}$ of the Lévy process defined by $\psi_{S}(\xi)$; the sub-symbol of the process \ref{eq:sde1} around $w_{\mathcal{D}}$. The choice of the SDE \ref{eq:sde1} imposes some structure on $\psi_{\mathcal{D}}$, which lets us express $\beta_{\mathcal{D}}$ in a simpler form. This helps us in estimating the BG index for a general Lévy process. As shown by Simsekli et al., there is a layer-wise variation of the tail-index of the gradient noise in a DNN-based multi-layer neural network\citep{simsekli2019tail}. Thus, for our STDP model we assume that around the local minimum $w_{\mathcal{D}}$, the dynamics of STDP will be similar to the Lévy motion with frozen coefficients: $\Sigma_{2}\left(w_{S}\right) \mathrm{L}^{\alpha\left(w_{\mathcal{D}}\right)}$. Also assuming the coordinates corresponding to the same layer $l$ have the same tail-index $\alpha_{l}$ around $w_{\mathcal{D}}$, the BG index can be analytically computed as $\beta_{\mathcal{D}}=\max _{l} \alpha_{l} \in(0,2]$  \citep{meerschaert2005dimension}. We note here that $\operatorname{dim}_{H} \mathcal{W}_{\mathcal{D}}$ and thus, $\beta_{\mathcal{D}}$ determines the order of the generalization error. Using this simplification, we can easily compute $\beta_{\mathcal{D}}$, by first estimating each $\alpha_{l}$ by using the estimator proposed by Mohammadi et al. \citep{mohammadi2015estimating}, that can efficiently estimate $\alpha_{l}$ by using multiple iterations of the STDP learning process.

}

As explained by Simsekli et al. \citep{simsekli2020hausdorff}, due to the decomposability property of each dataset $\mathcal{D}$, the stochastic process for the synaptic weights given by $W^{(\mathcal{D})}(t)$ behaves like a Lévy motion around a local point $w_0$. Because of this locally regular behavior, the Hausdorff dimension can be bounded by the Blumenthal-Getoor (BG) index ~\citep{blumenthal1960some}, which in turn depends on the tail behavior of the Lévy process. Thus, in summary, we can use the BG-index as a bound for the Hausdorff dimension of the trajectories from the STDP learning process. Now, as the Hausdorff dimension is a measure of the generalization error and is also controlled by the tail behavior of the process, heavier-tails imply less generalization error \citep{simsekli2020hausdorff}, \citep{simsekli2020fractional}.

\begin{figure}
    \centering
    \includegraphics[width = \textwidth]{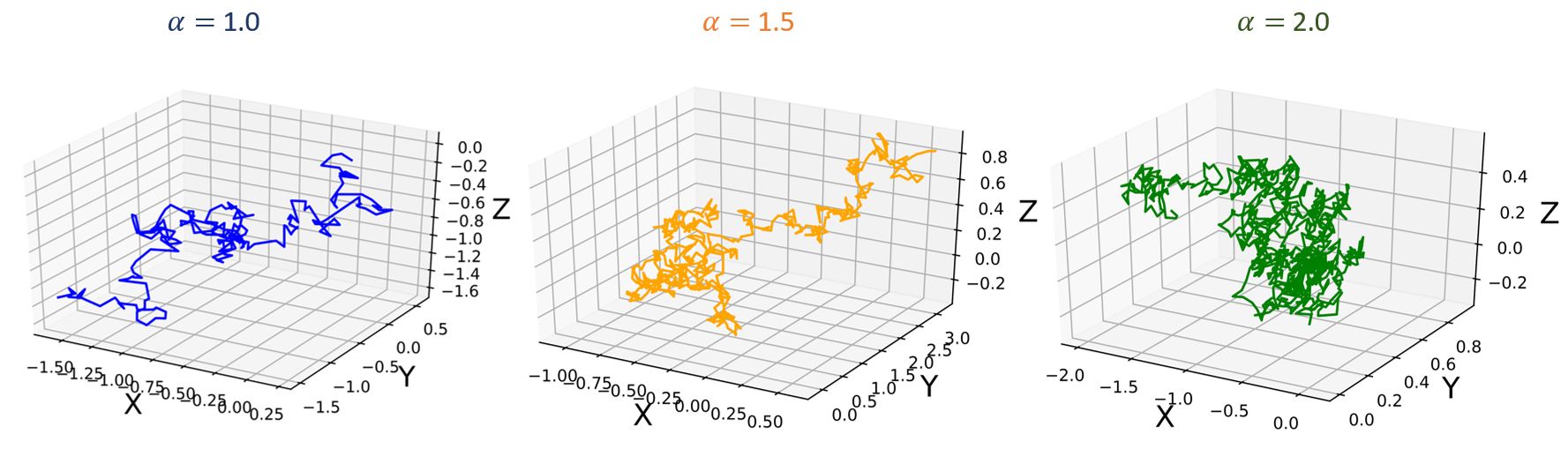}
    \caption{\hlt{Plot showing the trajectories of the $\alpha-$stable Lévy process $L^{\alpha}_{t}$ for varying values of $\alpha$}}
    \label{fig:alphas_traj}
\end{figure}

\begin{figure}
    \centering
    \includegraphics[width = 0.55\textwidth]{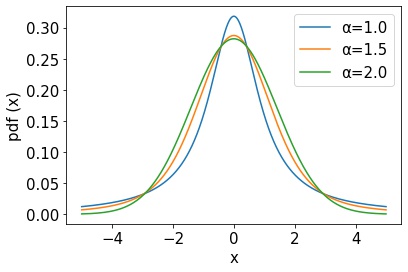}
    \caption{\hlt{Figure showing the probability density functions of the $\alpha-$stable Lévy process $L^{\alpha}_{t}$ for varying values of $\alpha$}}
    \label{fig:pdf_alpha}
\end{figure}

To further demonstrate the heavy-tailed behavior of the Ornstein-Uhlenbeck process (a type of $\alpha-$stable Lévy process $L^{\alpha}_{t}$) that characterizes the STDP learning mechanism, we plot its trajectories and their corresponding pdf distributions. We plot these for varying values of the stability factor of the Lévy process $L^{\alpha}_{t}$, $\alpha$. We hence also plotted the probability density function of the Lévy processes to show the heavy-tailed nature of the Lévy trajectories as the tail index $\alpha$ decreases.
Figure \ref{fig:alphas_traj} shows a continuous-time random walk performed using the O-U random process in 3D space. In the Figure, X, Y, Z are random variables for the $\alpha$-stable distributions generated using the O-U process. Fig. \ref{fig:pdf_alpha}, shows the corresponding probability density function of the O-U process for  varying values of $\alpha$ corresponding to the different trajectories shown in Fig. \ref{fig:alphas_traj}. From Figs. \ref{fig:alphas_traj},\ref{fig:pdf_alpha}, that as the O-U process becomes heavier-tailed (i.e., $\alpha$ decreases), and the Hausdorff dimension $\texttt{dim}_{\text{H}}$ gets smaller. 
}

\subsection{Optimal Hyperparameter Selection}
\hlt{Using the Hausdorff Dimension as a metric for the generalizability of the learning process, we formulate an optimization process that selects the hyperparameters of the STDP process to improve the generalizability of the models. The Hausdorff dimension is a function of the hyperparameters of the STDP learning process. Thus, we formulate an optimization problem to select the optimal hyperparameters of the STDP using the Hausdorff dimension of the STDP learning process as the optimization function. Now, since the BG-index is the upper bound of the Hausdorff dimension, as discussed earlier, we in turn aim to optimize the BG-index of the STDP stochastic process. The optimization problem aims to get the optimal set of hyperparameters of the STDP process that can give a more generalizable model without looking at the test set data.}
Now, given an STDP process, we aim to tune its hyperparameters to search for a  more generalizable model.
Let us define \hlt{$\boldsymbol{\lambda} := \{\lambda_1, \ldots, \lambda_N\}$ where $\boldsymbol{\lambda}$ is the set of $N$ hyperparameters of the STDP process, $\lambda_1, \ldots, \lambda_N$. Let $\Lambda_i$ denote the domain of the hyperparameter $\lambda_i$.} The hyperparameter space of the algorithm is thus defined as \hlt{$\boldsymbol{\Lambda} = \Lambda_1 \times \ldots \times \Lambda_N$}. Now, we aim to design the optimization problem to minimize the Hausdorff Dimension of the learning trajectory for the STDP process. This is calculated over the last \hlt{training iteration} of the model, assuming that it reaches near the local minima.
When trained with \hlt{$\boldsymbol{\lambda} \in \boldsymbol{\Lambda}$} on the training data $\mathcal{D}_{\textbf{train}}$, we denote the algorithm's Hausdorff dimension as $\operatorname{dim}_{\mathrm{H}} G(\boldsymbol{\lambda}; \mathcal{D}_{\text{train}})$. Thus, using K-fold cross validation, the hyperparameter optimization problem for a given dataset $\mathcal{D}$ is to given as follows:
\begin{equation}
\label{eq:optim}
     \boldsymbol{\lambda_{s}}= \underset{\boldsymbol{\lambda} \in \boldsymbol{\Lambda}}{\arg \min }  \frac{1}{K} \sum_{i=1}^{K} \operatorname{dim}_{\mathrm{H}} G(\boldsymbol{\lambda}; \mathcal{D}_{\text{train}})
\end{equation}
We choose the Sequential Model-based Bayesian Optimization (SMBO) technique for this problem ~\citep{feurer2015initializing}.
SMBO constructs a probabilistic model $\mathcal{M}$ of $f =  \operatorname{dim}_{\mathrm{H}} G$ based on point evaluations of $f$ and any available prior information. \hlt{It then} uses that model to select subsequent configurations $\boldsymbol{\lambda}$ to evaluate. \hlt{Given a set of hyperparameters $\boldsymbol{\lambda}$ for an STDP learning process $G$, we define the point functional evaluation as the calculation of the BG index of $G$ with the hyperparameters $\boldsymbol{\lambda}$. The BG index gives an upper bound on the Hausdorff dimension of the learning process. } In order to select its next hyperparameter configuration $\boldsymbol{\lambda}$ using model $\mathcal{M},$ SMBO uses an acquisition function $a_{\mathcal{M}}: \boldsymbol{\lambda} \rightarrow \mathbb{R},$ which uses the predictive distribution of model $\mathcal{M}$ at arbitrary hyperparameter configurations $\boldsymbol{\lambda} \in \boldsymbol{\Lambda}$. This function is then maximized over $\boldsymbol{\Lambda}$ to select the most useful configuration $\boldsymbol{\lambda}$ to evaluate next. There exists a wide range of acquisition functions ~\citep{mockus1978application},  all of whom aim to trade-off between exploitation and exploration. The acquisition function tries to balance between locally optimizing hyperparameters in regions known to perform well and trying hyperparameters in a relatively unexplored region of the space.

In this paper, for the acquisition function, we use the expected improvement ~\citep{mockus1978application} over the best previously-observed function value $f_{\min }$ attainable at a hyperparameter configuration $\boldsymbol{\lambda}$  where expectations are taken over predictions with the current model $\mathcal{M}$:
\begin{equation}
    a(\boldsymbol{\lambda}, \mathcal{M})=\int_{-\infty}^{f_{\min }} \max \left\{f_{\min }-f, 0\right\} \cdot p_{\mathcal{M}}(f \mid \boldsymbol{\lambda}) df
\end{equation}

\section{RESULTS}

\subsection{Experimental Setup}

We empirically study the generalization properties of the STDP process by designing an SNN \hlt{with 6400 learning neurons} for hand-written digit classification using the MNIST dataset. The MNIST dataset contains $60,000$training examples and $10,000$ test examples of $28\times28$ pixel images of the digits $0–9$. \hlt{It must be noted here that the images from the ten classes in the MNIST dataset are randomized so that there is a reinforcement of the features learned by the network.}

\begin{figure}
    \centering
    \includegraphics[width = 0.5\columnwidth]{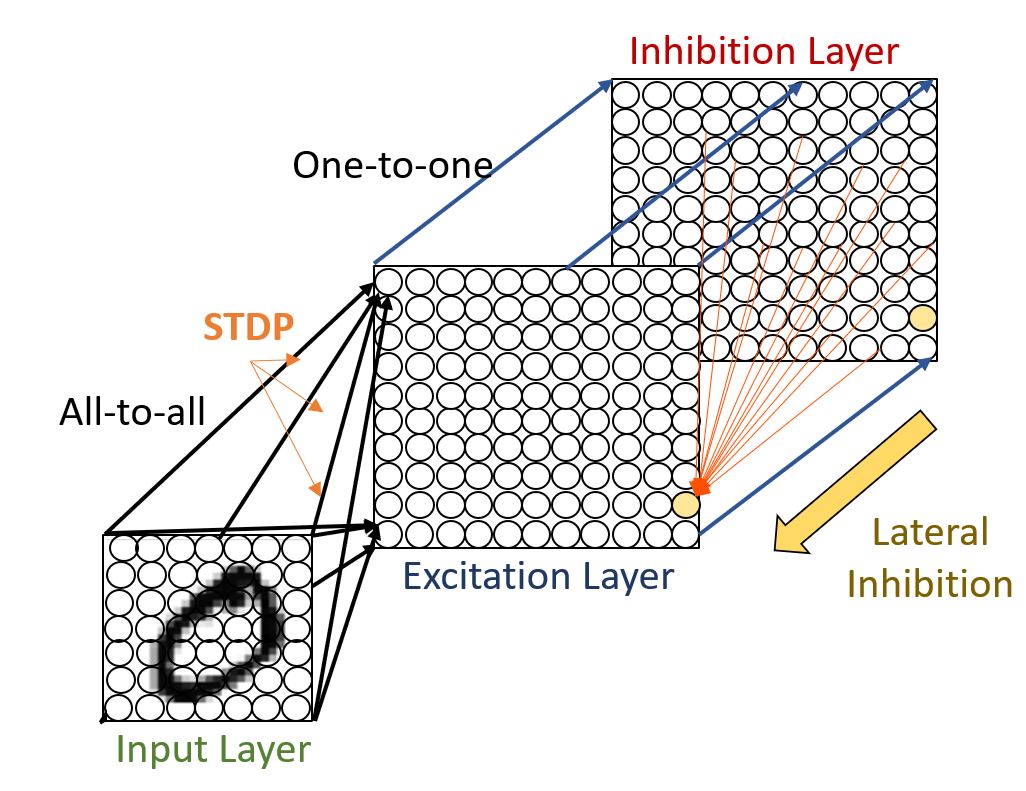}
    \caption{\hlt{The intensity values of the MNIST image are converted to Poisson-spike trains. The firing rates of the Poisson point process are proportional to the intensity of the corresponding pixel. These spike trains are fed as input in an all-to-all fashion to excitatory neurons. In Fig., the black shaded area from the input to the excitatory layer shows the input connections to one specific excitatory example neuron. The red shaded area denotes all connections from one inhibitory neuron to the excitatory neurons. While the excitatory neurons are connected to inhibitory neurons via one-to-one connection, each of the inhibitory neurons is connected to all excitatory ones, except for the one it receives a connection from.}}
    \label{fig:diehl}
\end{figure}

\textbf{Architecture}. We use a two-layer SNN architecture \hlt{as done by the} implementation of Diehl et al., ~\citep{diehl2015unsupervised} as shown in Figure \ref{fig:diehl}. The first layer is the input layer, containing $28 \times 28$ neurons with one neuron per image pixel. The second layer is the processing layer, with an equal number of excitatory and inhibitory neurons. \hlt{The excitatory neurons of the second layer are connected in a one-to-one fashion to inhibitory neurons such that each spike in an excitatory neuron will trigger a spike in its corresponding inhibitory neuron. Again, each of the inhibitory neurons is connected to all excitatory ones, except for the one from which it receives a connection. This connectivity provides lateral inhibition and leads to competition among excitatory neurons. There is a balance between the ratio of inhibitory and excitatory synaptic conductance to ensure the correct strength of lateral inhibition. For a weak lateral inhibition, the conductance will not have any influence while an extremely strong signal would ensue that one dominant neuron suppresses the other ones.

\par The task for the network is to learn a representation of the dataset on the synapses connecting the input layer neurons to the excitatory layer neurons. The excitatory-inhibitory connectivity pattern creates competition between the excitatory neurons. This allows individual neurons to learn unique filters where the single most spiked neuron in each iteration updates its synapse weights to match the current input digit using the chosen STDP rule. Increasing the number of neurons allows the network to memorize more examples from the training data and recognize similar patterns during the test phase.
}

\hlt{\textbf{Homeostasis}. The inhomogeneity of the input leads to different firing rates of the excitatory neurons, and lateral inhibition further increases this difference. However, all neurons should have approximately equal firing rates to prevent single neurons from dominating the response pattern and to ensure that the receptive fields of the neurons differentiate. To achieve this, we employ an adaptive membrane threshold resembling intrinsic plasticity \citep{zhang2003other}. Specifically, each excitatory neuron's membrane threshold is not only determined by $v_{\text{thresh}}$ but by the sum $v_{\text{thresh}} + \theta$, where $\theta$ is increased every time the neuron fires and is exponentially decaying \citep{querlioz2013immunity}. Therefore, the more a neuron fires, the higher will be its membrane threshold and in turn, the neuron requires more input to a spike soon. Using this mechanism, the firing rate of the neurons is limited because the conductance-based synapse model limits the maximum membrane potential to the excitatory reversal potential $E_{\text{exc}}$, i.e., once the neuron membrane threshold is close to $E_{\text{exc}}$ (or higher) it will fire less often (or even stop firing completely) until $\theta$ decreases sufficiently.
}

\hlt{\textbf{Input Encoding}. The input image is converted to a Poisson spike train with firing rates proportional to the intensity of the corresponding pixel. This spike train is then presented to the network in an all-to-all fashion for 350 ms as shown in Fig. \ref{fig:diehl}. } \hlt{ The maximum pixel intensity of 255 is divided by 4, resulting in input firing rates between 0 and 63.75 Hz. Additionally, if the excitatory neurons in the second layer fire less than five spikes within 350 ms, the maximum input firing rate is increased by 32 Hz and the example is presented again for 350 ms. This process is repeated until at least five spikes have been fired during the entire time the particular example was presented.}

\textbf{Training and STDP Dynamics Analysis}. To train the network, we present digits from the MNIST training set to the network. It is to be noted that, before presenting a new image, no input to any variable of any neuron is given for a time interval of 150 ms. This is done to decay to their resting values. All the synaptic weights from input neurons to excitatory neurons are learned using the STDP learning process as described in Sec. \ref{sec:stdp}. To improve simulation speed, the weight dynamics are computed using synaptic traces such that every time a \hlt{pre-synaptic} spike $x_{pre}$ arrives at the synapse, the trace is increased by 1 ~\citep{morrison2007spike}. Otherwise, $x_{pre}$ decays exponentially. When a \hlt{post-synaptic} spike arrives at the synapse the weight change $\Delta w$ is calculated based on the pre-synaptic trace as described in section \ref{sec:stdp}. \hlt{To evaluate the model, we divide the training set into 100 divisions of 600 images each and check the model performance after each such batch is trained using the STDP learning model. In the remainder of the paper, we call this evaluation of the model after 600 images as one iteration.}

\textbf{Inference}. After the training process is done, the learning rate is set to zero and each neuron’s spiking threshold is fixed.  A class is assigned to each neuron based on its highest response to the ten classes of digits over one presentation of the training set. This is the first time labels are used in the learning algorithm, which makes it an unsupervised learning method. \hlt{The response of the class-assigned neurons is used to predict the digit. It is determined by taking the mean of each neuron response for every class and selecting the class with the maximum average firing rate. These predictions are then used to measure the classification accuracy of the network on the MNIST test set}

\begin{table}[]
\centering
\caption{\hlt{Table showing the set of hyperparameters for various STDP processes}}
\hlt{
\label{tab:hypers}
\begin{tabular}{|c|c|c|c|}
\hline
\textbf{Hyperparameter} & \textbf{logSTDP} & \textbf{addSTDP} & \textbf{multSTDP} \\ \hline
Synaptic Delay & 0.75ms & 0.75ms & 0.75ms \\ \hline
Synaptic epsp $\tau_A$ & 1ms & 1ms & 1ms \\ \hline
Synaptic epsp $\tau_B$ & 5ms & 5ms & 5ms \\ \hline
Number of correlated pools & 4 & 4 & 4 \\ \hline
Number of neurons per pool & 50 & 50 & 50 \\ \hline
Spiking Rate of inputs & 10Hz & 10Hz & 10Hz \\ \hline
Learning rate ($\eta$) & $0.0002$ & $0.0002$ & $0.0002$ \\ \hline
STDP Apre (LTP) time constant & 17ms & 17ms & 17ms \\ \hline
STDP Apre (LTD) time constant & 34ms & 34ms & 34ms \\ \hline
Increase in $A_{\text{pre}}$ (LTP), on pre-spikes $A_{\text{pre}0}$ & 1.0 & 1.0 & 1.0 \\ \hline
Increase in $A_{\text{post}}$ (LTD), on post-spikes $A_{\text{post}0}$ & 0.5 & 0.55 & 100 \\ \hline
LTD Curvature Factor ($\alpha$) & 5 & N/A & N/A \\ \hline
Exponential LTP Decay factor ($\beta$) & 50 & N/A & N/A \\ \hline
Threshold Fixed-point weight ($W_0$) & 0.006 & N/A & N/A  \\ \hline
\end{tabular}%
}
\end{table}

\textbf{Computation of Generalization Error and Hausdorff Dimension}. We empirically study the generalization behavior of STDP trained SNNs.  We vary the hyperparameters of the STDP learning process which controls the LTP/LTD dynamics of the STDP learning algorithm. Table \ref{tab:hypers} shows the hyperparameters for various STDP processes. We trained all the models for 100 \hlt{training iterations}. \hlt{In this paper, we consider the synaptic weight update to follow a Lévy process, i.e., continuous with discrete jumps similar to the formulation of Stein et al. \citep{stein1965theoretical} and Richardson et al. \citep{richardson2010firing}.} As discussed in Section \ref{sub:our-approach}, the generalizability can be measured using the Hausdorff dimension which is bounded by BG-index.

\hlt{Therefore, the BG-index is computed in the last iteration when all the neurons have learned the input representations. 
We also compute the generalization error as the difference between the training and test accuracy. we study the relations between BG-index, generalization error, testing accuracy, and convergence behavior of the networks. }

\subsection{Analysis of Generalizability of STDP Processes}
\label{sec:hausdorff}
\textbf{Impact of Scaling Functions}.
Kubota et al. showed that the scaling functions play a vital role in controlling the LTP/LTD dynamic of the STDP learning method ~\citep{kubota2009modulation}. In this subsection, we evaluate the impact of scaling functions (i.e. $a_\pm$ in the equation \ref{eq:window}) on the generalizability properties of the STDP methods.
We define the ratio of the post-synaptic scaling function to the pre-synaptic one (i.e. $c_+/c_-$ in add-, mult-, and log- STDP equations), hereafter referred to as the \hlt{scaling function ratio (SFR)}, as our variable.
Kubota et al. has shown that the learning behavior is best when this SFR lies between the range of 0.9 to 1.5. Hence, we also modulate the SFR within this set interval.
Table \ref{tab:log_scaling} shows the impact of scaling function on Hausdorff dimension (measured using BG-index), generalization error, and testing accuracy. We observe that a smaller SFR leads to a lower Hausdorff dimension and a lower generalization error, while a higher ratio infers a less generalizable model. However, a higher SFR marginally increases the testing accuracy.  The analysis shows confirms that a higher Hausdorff dimension (i.e. a higher BG-index) corresponds to a higher generalization error, as discussed in section \ref{sub:our-approach}, justifying the use of BG-index as a measure of the generalization error.

\begin{table*}[]
\centering
\caption{Impact of the Post-synaptic to Pre-synaptic Scaling Functions Ratio on Generalization}
\label{tab:log_scaling}
\resizebox{\textwidth}{!}{%
\begin{tabular}{|c|c|c|c|c|c|c|c|c|c|}
\hline
$\frac{c_+}{c_-}$ & \multicolumn{3}{c|}{\textbf{log-stdp}} & \multicolumn{3}{c|}{\textbf{add-stdp}} & \multicolumn{3}{c|}{\textbf{mult-stdp}} \\ \cline{2-10} 
 & \textbf{\begin{tabular}[c]{@{}c@{}}BG\\ Index\end{tabular}} & \textbf{\begin{tabular}[c]{@{}c@{}}Generalization\\ Error\end{tabular}} & \textbf{\begin{tabular}[c]{@{}c@{}} Testing\\ Accuracy\end{tabular}} & \textbf{\begin{tabular}[c]{@{}c@{}}BG\\ Index\end{tabular}} & \textbf{\begin{tabular}[c]{@{}c@{}}Generalization\\ Error\end{tabular}} & \textbf{\begin{tabular}[c]{@{}c@{}}Testing\\ Accuracy\end{tabular}} & \textbf{\begin{tabular}[c]{@{}c@{}}BG \\ Index\end{tabular}} & \textbf{\begin{tabular}[c]{@{}c@{}}Generalization\\ Error\end{tabular}} & \textbf{\begin{tabular}[c]{@{}c@{}} Testing\\ Accuracy\end{tabular}} \\ \hline
2.1 & 1.352 & 6.8 & 89.92 & 1.969 & 9.7 & 88.17 & 1.824 & 8.1 & 89.26 \\ \hline
1.7 & 1.294 & 6.2 & 89.98 & 1.911 & 9.3 & 88.12 & 1.797 & 7.6 & 89.15 \\ \hline
1.2 & 1.209 & 5.9 & 89.79 & 1.875 & 8.9 & 88.09 & 1.702 & 7.0 & 88.99 \\ \hline
0.9 &  1.174 & 5.7 & 89.26 & 1.799 & 8.6 & 88.10 & 1.633 & 6.5 & 88.87\\ \hline
\end{tabular}%
}
\end{table*}

\begin{figure}
    \centering
    \includegraphics[width = \textwidth]{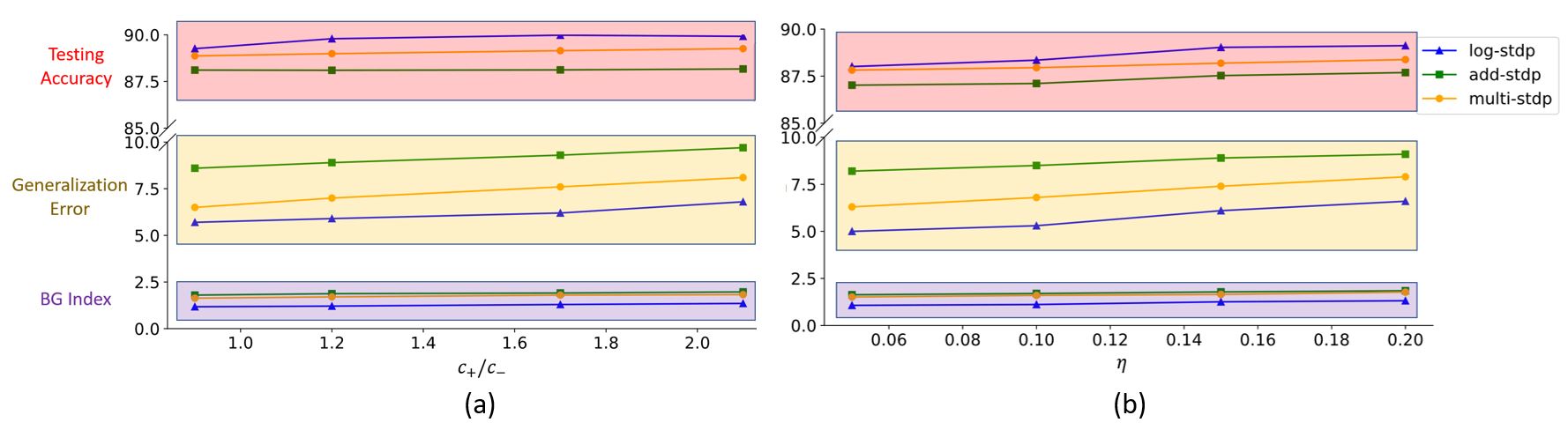}
    \caption{\hlt{(a) Plots for the impact of the scaling function ratios on generalization (results shown in Table \ref{tab:log_scaling}) (b) Plots for the impact of the learning rates on generalization (results shown in Table \ref{tab:log_lr}) }}
    \label{fig:table_plots}
\end{figure}

\textbf{Impact of the Learning Rate}. One of the major parameters that control the weight dynamics of the STDP processes is the learning rate i.e. the variable $\eta$ in equation \ref{eq:synaptic_weight}. In this subsection, we evaluate the effect of the learning rate on the generalizability of the STDP process. We have summarized the results in Table \ref{tab:log_lr}. We observe that a larger learning rate converges to a more generalizable solution. This can be attributed to the fact that a higher learning rate inhibits convergence to sharper minimas; rather facilitates convergence to a flatter one resulting in a more generalizable solution. We also observe the monotonic relation between the BG-index and the generalization error.

\begin{table*}[]
\centering
\caption{The Impact of Learning Rate on the Generalization Error}
\label{tab:log_lr}
\resizebox{\textwidth}{!}{%
\begin{tabular}{|c|c|c|c|c|c|c|c|c|c|}
\hline
$\eta$ & \multicolumn{3}{c|}{\textbf{log-stdp}} & \multicolumn{3}{c|}{\textbf{add-stdp}} & \multicolumn{3}{c|}{\textbf{mult-stdp}} \\ \cline{2-10} 
 & \textbf{\begin{tabular}[c]{@{}c@{}}BG\\ Index\end{tabular}} & \textbf{\begin{tabular}[c]{@{}c@{}}Generalization\\ Error\end{tabular}} & \textbf{\begin{tabular}[c]{@{}c@{}}Testing\\ Accuracy\end{tabular}} & \textbf{\begin{tabular}[c]{@{}c@{}}BG\\ Index\end{tabular}} & \textbf{\begin{tabular}[c]{@{}c@{}}Generalization\\ Error\end{tabular}} & \textbf{\begin{tabular}[c]{@{}c@{}}Testing\\ Accuracy\end{tabular}} & \textbf{\begin{tabular}[c]{@{}c@{}}BG \\ Index\end{tabular}} & \textbf{\begin{tabular}[c]{@{}c@{}}Generalization\\ Error\end{tabular}} & \textbf{\begin{tabular}[c]{@{}c@{}}Testing\\ Accuracy\end{tabular}} \\ \hline
0.2 & 1.312 & 6.6 & 89.12 & 1.844 & 9.1 & 87.69 & 1.769 & 7.9 & 88.38 \\ \hline
0.15 & 1.255 & 6.1 & 89.03 & 1.783 & 8.9 & 87.53 & 1.648 & 7.4 & 88.19 \\ \hline
0.1 & 1.112 & 5.3 & 88.35 & 1.698 & 8.5 & 87.11 & 1.596 & 6.8 & 87.95 \\ \hline
0.05 & 1.068 & 5.0 & 88.01 & 1.632 & 8.2 & 87.02 & 1.512 & 6.3 & 87.82 \\ \hline
\end{tabular}%
}
\end{table*}

\textbf{Impact of the STDP models on Generalizability}. 
In this section, we compare the three different STDP models, namely, add-STDP, mult-STDP, and log-STDP to its generalization abilities with changing SFR \hlt{(scaling function ratio)}  and learning rate. The results are summarized in Tables \ref{tab:log_scaling}, \ref{tab:log_lr}. In all the above cases we see that the log-STDP process has a significantly lower generalization error compared to the other two STDP methods. 
The difference between the generalizability of various STDP models comes from the nature of the stochastic distribution of weights generated by different models.

Gilson et al. ~\citep{gilson2011stability} has discussed that add-STDP ~\citep{gutig2003learning} can rapidly and efficiently select synaptic pathways by splitting synaptic weights into a bimodal distribution of weak and strong synapses. However, the stability of the weight distribution requires hard bounds due to the resulting unstable weight dynamics. In contrast in mult-STDP ~\citep{rubin2001equilibrium}, weight-dependent update rules can generate stable unimodal distributions. However, mult-STDP weakens the competition among synapses leading to only weakly skewed weight distributions. The probability distributions of the three different STDP models are shown in Fig. \ref{fig:stdp_comp}. On the other hand, log-STDP proposed by Gilson et al.  ~\citep{gilson2011stability} bypass these problems by using a weight-dependent update rule while does not make the other synapses weak as in mult-STDP. The log-STDP results in a log-normal solution of the synaptic weight distribution as discussed by Gilson et al. ~\citep{gilson2011stability}. A log-normal solution has a heavier tail and thus a \hlt{smaller} Hausdorff dimension leading to a lower generalization error. A detailed comparison of the weight distributions of the three types of STDP processes can be found in the paper by Gilson et al. ~\citep{gilson2011stability}.
\hlt{
We further evaluated the training loss for iterations for the different STDP models. The results are plotted in Fig. \ref{fig:log_mult_add_sfr}. From the figures we see that the log-STDP outperforms the add-STDP and the mult-STDP in terms of training loss for either case, thus demonstrating the performance of  

}

 \begin{figure}
     \centering
      \includegraphics[width = 0.6\columnwidth]{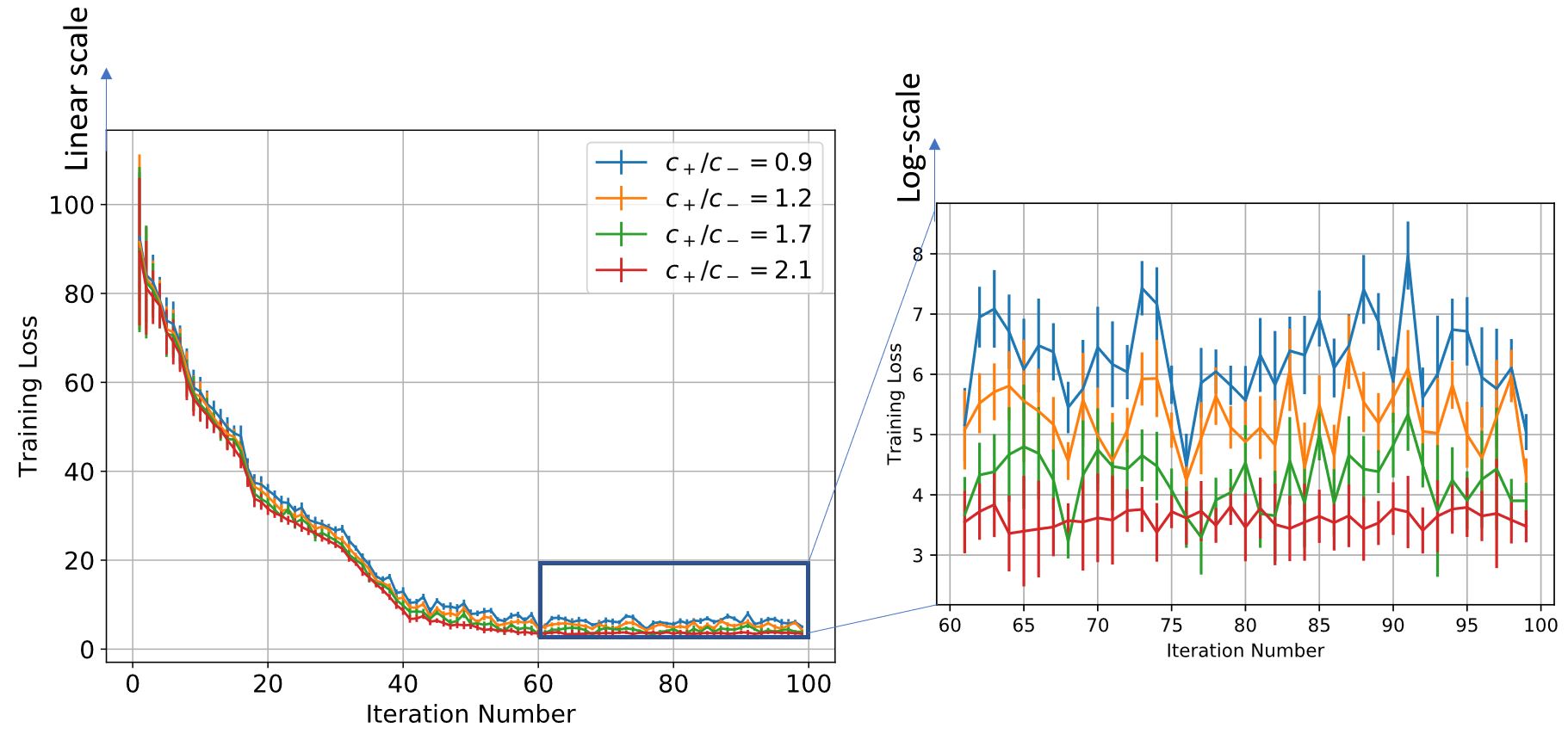}
     \caption{\hlt{Figure showing the change in training loss with iterations for varying scaling function ratios for the log-STDP learning process}}
     \label{fig:haus_log_sc}
 \end{figure}

 \begin{figure}
     \centering
    \includegraphics[width = 0.6\columnwidth]{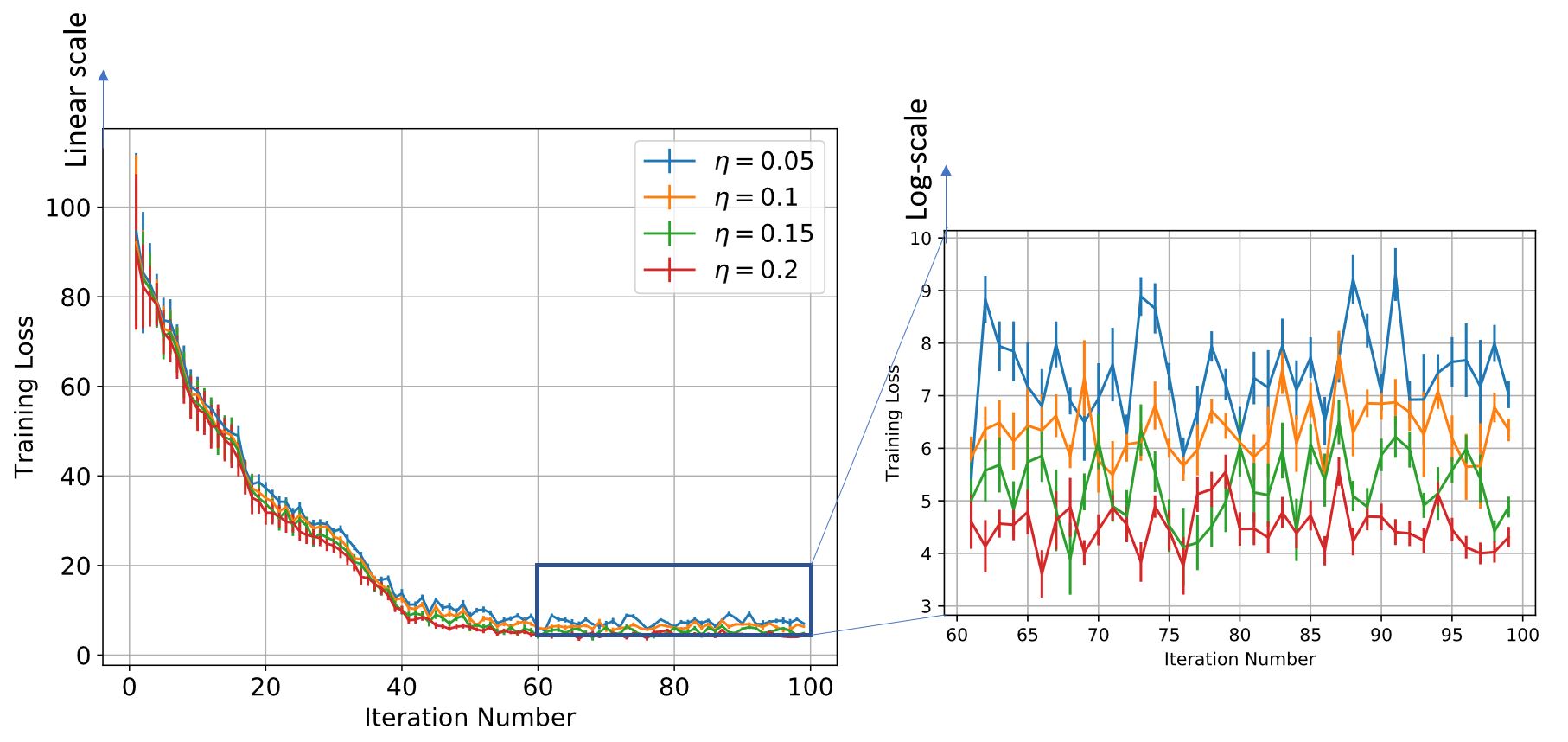}
     \caption{\hlt{Figure showing the change in training loss with iterations for varying learning rates for the log-STDP learning process}}
     \label{fig:haus_log_lr}
 \end{figure}

 \begin{figure}
     \centering
      \includegraphics[width =\columnwidth]{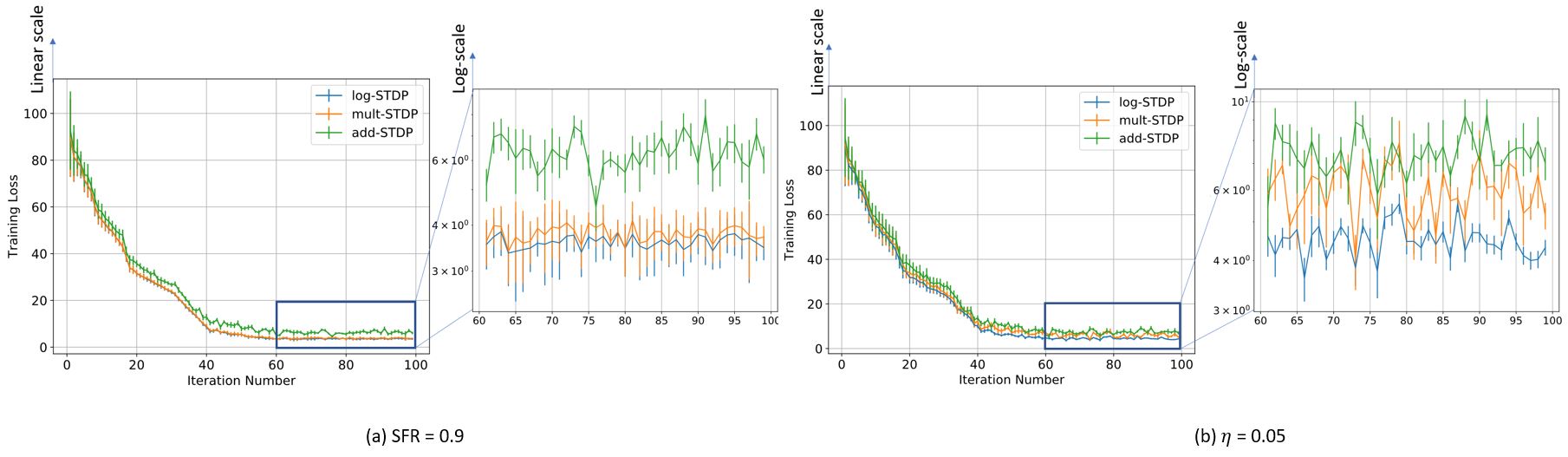}
     \caption{\hlt{Figure showing the variation in the training loss with increasing iterations for different types of STDP models keeping (a) SFR= 0.9 and (b) $\eta = 0.05$ }}
     \label{fig:log_mult_add_sfr}
 \end{figure}

\hlt{
\textbf{Impact on Different Datasets:}
To demonstrate the generalizability of the STDP models, we also tested its performance on Fashion-MNIST \citep{xiao2017fashion} which is an MNIST-like fashion product dataset with 10 classes. Fashion-MNIST shares the same image size and structure of the training and testing splits as MNIST but is considered more realistic as its images are generated from front look thumbnail images of fashion products on Zalando’s website via a series of conversions. Therefore, Fashion-MNIST poses a more challenging classification task than MNIST. We preprocessed the data by normalizing the sum of a single sample gray value because of the high variance among examples. The results for SFR $c_+/c_- = 0.9$ and learning rate $\eta = 0.05$ is shown in Table \ref{tab:fashion}. We see that as seen in MNIST datasets, for the Fashion-MNIST also, the log-STDP method has a lower generalization error corresponding to a lower BG Index.

\begin{table}[]
\centering
\hlt{
\resizebox{\textwidth}{!}{%
\begin{tabular}{|c|c|c|c|c|c|c|c|c|c|}
\hline
\multirow{2}{*}{}    & \multicolumn{3}{c|}{\textbf{log-STDP}}                                                                                                                                                                         & \multicolumn{3}{c|}{\textbf{add-STDP}}                                                                                                                                                                         & \multicolumn{3}{c|}{\textbf{mult-STDP}}                                                                                                                                                                        \\ \cline{2-10} 
                     & \textbf{\begin{tabular}[c]{@{}c@{}}BG \\ Index\end{tabular}} & \textbf{\begin{tabular}[c]{@{}c@{}}Generalization \\ Error\end{tabular}} & \textbf{\begin{tabular}[c]{@{}c@{}}Testing \\ Accuracy\end{tabular}} & \textbf{\begin{tabular}[c]{@{}c@{}}BG\\  Index\end{tabular}} & \textbf{\begin{tabular}[c]{@{}c@{}}Generalization \\ Error\end{tabular}} & \textbf{\begin{tabular}[c]{@{}c@{}}Testing \\ Accuracy\end{tabular}} & \textbf{\begin{tabular}[c]{@{}c@{}}BG \\ Index\end{tabular}} & \textbf{\begin{tabular}[c]{@{}c@{}}Generalization \\ Error\end{tabular}} & \textbf{\begin{tabular}[c]{@{}c@{}}Testing \\ Accuracy\end{tabular}} \\ \hline
\textbf{c+/c\_ =0.9} & 1.322                                                        & 10.02                                                                    & 82.43                                                                & 1.788                                                        & 13.87                                                                    & 79.16                                                                & 1.573                                                        & 11.38                                                                    & 81.92                                                                \\ \hline
$\mathbf{\eta=0.05}$      & 1.204                                                        & 9.93                                                                     & 81.75                                                                & 1.692                                                        & 11.73                                                                    & 79.76                                                                & 1.489                                                        & 10.11                                                                    & 80.35                                                                \\ \hline
\end{tabular}
}
\caption{\hlt{Table Showing comparison of the STDP models on the Fashion-MNIST dataset}}
\label{tab:fashion}
}
\end{table}

}

\subsection{Generalizability vs Trainability Tradeoff}

In this section, we study the relations between the generalizability and trainability of a learning model. For the sake of brevity, we only focus on the log-STDP process as it has shown better generalizability compared to add-STDP and mult-STDP. 
 
We plot the training loss as a function of the time evolution of the synaptic weights trained with the STDP learning method. \hlt{ Since STDP is an online unsupervised learning algorithm, there is no formal concept of training loss. So, to evaluate the performance of the model, we define the training loss as follows:
We divide the MNIST dataset into 100 divisions, with each division consisting of 600 images. We evaluate the model after training the model on each subset of the full training dataset and this is considered as one training iteration. We train the SNN model using STDP with this limited training dataset. After the training is done, we set the learning rate to zero and fix each neuron's spiking threshold. Then, the image of each class is given as input to the network, and the total spike count of all the neurons that have learned that class is calculated. Hence, the spike counts are normalized by dividing the number of spikes by the maximum number of spike counts and the cross-entropy loss was calculated across all the classes. This is defined as the training loss. To show the confidence interval of each training iteration, we evaluated each of the partially trained models on the test dataset 5 times, randomizing the order of images for each of the test runs. We see from Figure 6, that initially, as the model was trained with fewer images, the variance in training loss was high demonstrating a low confidence interval. However, as the model is trained on a larger training set, the variance decreases as expected.
}

Table 2 and Figure 4 show the training loss versus the number of iterations for the log-STDP process for various SFR. We see that the $SFR = c_+/c_- = 0.9$ shows a lower generalization error and almost similar testing accuracy, compared to the other SFRs. The results show that increasing the SFR increases the generalization error. If the pre-synaptic scaling function is stronger than the post-synaptic scaling function (i.e. $c_+/c_-$ is lower), it implies that the synaptic weights of the neurons gradually decay. \hlt{Since we have the images in the MNIST dataset randomized over the ten classes, the more important features which help in the generalization ability of the network over unknown data are reinforced so that the network does not forget these filters as shown by Panda et al. \cite{panda2017asp}.  Thus, the network only forgets the less important features and preserves the more important ones, hence making it more generalizable. Since the neuron forgets some features which would help to fit better into the current dataset, it affects its training/testing accuracy as can be seen in Tables 2,3. } Thus, the model learns the more important features and is essentially more generalizable.

\hlt{Note here that the training loss for the STDP processes all reach their convergence around iteration 60 - i.e., images after that added little information for the training of the model. The models here are not optimized and hence optimizing the hyperparameters can also help in reducing the number of images required for extracting enough information from the training dataset. Thus, if SFR is too high, training gets messed up since a neuron starts spiking for multiple patterns, in which case there is no learning. As the SFR value increases from 1, the SNN tends to memorize the training pattern and hence the generalization performance is poor. On the other hand, if when SFR is less than 1 but is close to 1, it is hard to memorize the training patterns as the STDP process tends to forget the patterns which are non-repeating, leading to better generalization.
}

On the other hand, if the post-synaptic scaling function is stronger than the pre-synaptic one (i.e. $c_+/c_-$ is higher), then the neurons tend to learn more than one pattern and respond to all of them. 
Similar results can be verified from Figure \ref{fig:haus_log_lr} where the learning rate was varied instead of the SFR. In this study as well we observe that a higher learning rate, although leads to faster convergence and lower training loss, leads to a less generalizable model.  
Hence, we empirically observe that hyperparameters of STDP models that lead to better generalizability can also make an SNN harder to train.

\subsection{Results of Hyperparameter Optimization}

In Section \ref{sec:hausdorff}, we empirically showed that the Hausdorff dimension is a good measure of the generalizability of the model and it can be very efficiently controlled using the hyperparameters of the STDP learning process. In this section, we show the application of our Bayesian optimization strategy to search for the optimal hyperparameters to increase the generalizability of an STDP-trained SNN model. 
For the sake of brevity, we demonstrate the application of Bayesian optimization on the log-STDP process. Table \ref{tab:opt_hyp} shows the set of hyperparameters that are optimized and their optimal values obtained by our approach. \hlt{ It should be clarified here that the hyper-parameters are not necessarily the absolute global optimum but a likely local optimum found in the optimization algorithm.} The optimized log-STDP model results in a training accuracy of 93.75\%, testing the accuracy of 90.49\%, and a BG Index of 0.718 for the MNIST dataset. 

\begin{figure}
    \centering
    \includegraphics[width = 0.6\columnwidth]{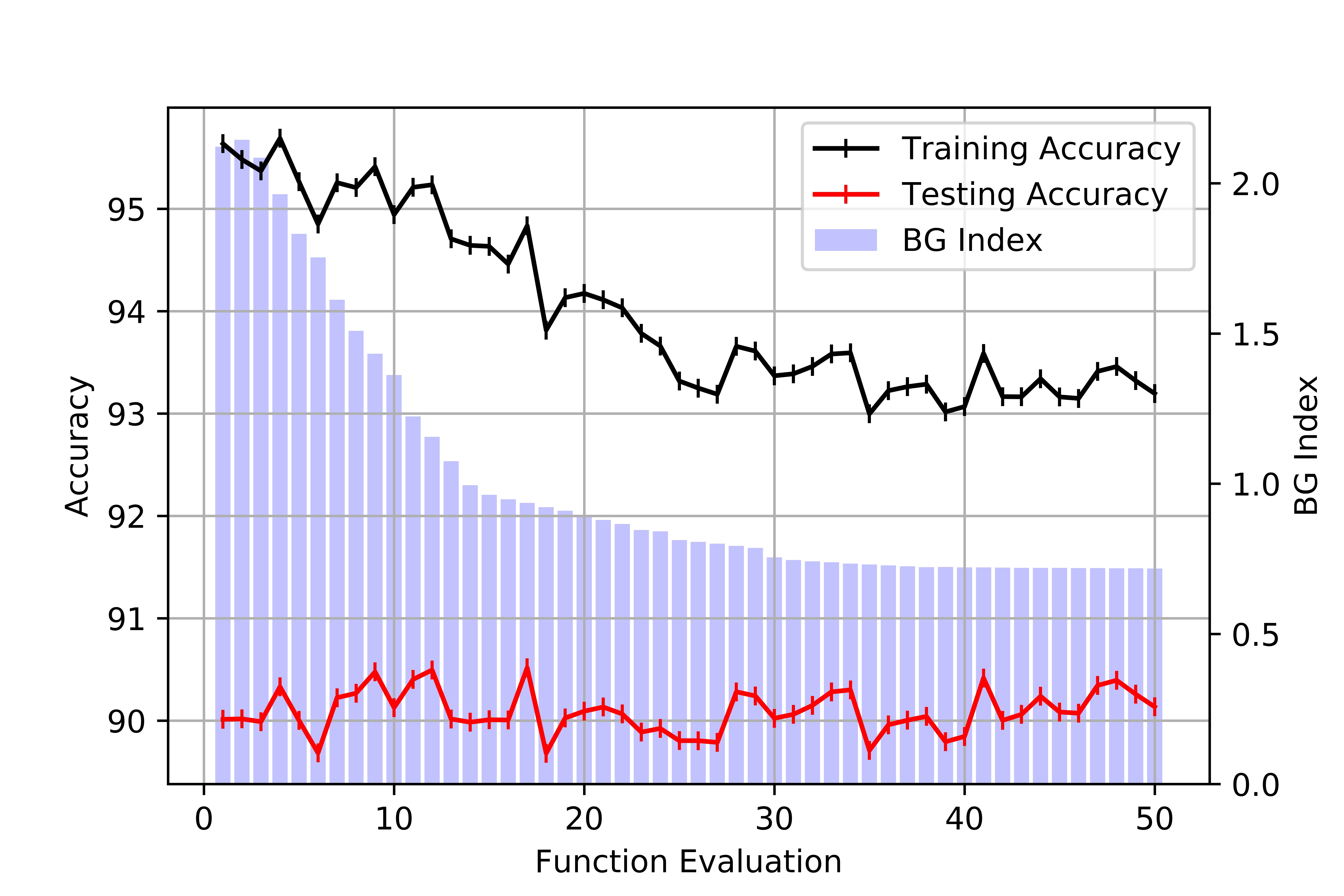}
    \caption{\hlt{Plot showing the change of BG Index and Training/Testing Accuracy over the Functional Evaluations of the Bayesian Optimization problem } }
    \label{fig:bayes_opt}
\end{figure}

We study the behavior of Bayesian optimization. Each iteration in the Bayesian optimization process corresponds to a different set of hyperparameters for the log-STDP model. \hlt{Each such iteration is called a functional evaluation. For each functional evaluation, the Bayesian Optimization trains the SNN with the corresponding set of hyperparameters of the log-STDP model and measures the BG-index of the weight dynamics of the trained-SNN.}  Figure \ref{fig:bayes_opt}(a) shows the change in the BG-Index as a function of a number of the function evaluations of the search process. It is to be noted here that at each functional evaluation, we train the network with the STDP learning rule with the chosen hyperparameters and estimate the Hausdorff dimension from the trained network.
We see that for the optimal set of hyperparameters, the BG Index converges to $0.7$. Figure \ref{fig:bayes_opt}(b) shows the corresponding training accuracy of the model with the change in iteration number. We see that the log-STDP configurations during Bayesian optimization that have a higher BG Index (i.e. a higher generalization error) also have has a higher training accuracy. These results further validate our observations on the generalizability vs trainability tradeoff for a log-STDP trained SNN. 

\begin{table}[]
\centering
\hlt{
\caption{\hlt{Table showing the set of hyperparameters used for the Bayesian optimization problem for the log-STDP process}}
\label{tab:opt_hyp}
\begin{tabular}{|c|c|c|c|c|c|}\hline
\multirow{2}{*}{\textbf{Hyperparameter}} & \multirow{2}{*}{\textbf{Domain}} & \multicolumn{2}{c|}{\textbf{logSTDP}}  & \multicolumn{2}{c|}{\textbf{add-STDP}} \\ \cline{3-6} 
                                         &                                  & \textbf{Before BO} & \textbf{After BO} & \textbf{Before BO} & \textbf{After BO} \\ \hline
\begin{tabular}[c]{@{}c@{}}Learning \\ rate ($\eta$)\end{tabular} & [0.05, 0.2] & 0.1 & 0.063 & 0.1 & 0.017\\ \hline
\begin{tabular}[c]{@{}c@{}} Variance of \\ Noise $\zeta$ ($\sigma$) \end{tabular} & [0.1, 1] & 0.5 & 0.581  & 0.5 & 0.632 \\ \hline
\begin{tabular}[c]{@{}c@{}} Degree of log-like\\ saturation ($\hlt{\mathcal{S}}$) \end{tabular} & $\mathbb{Z} \in [1, 10]$ & 3 & 5  & N/A & N/A\\ \hline
\begin{tabular}[c]{@{}c@{}}Exponential Decay \\ factor ($\hlt{\gamma}$) \end{tabular} & $\mathbb{Z} \in [10, 100]$  & 45 & 57    & N/A & N/A\\ \hline
\begin{tabular}[c]{@{}c@{}}Threshold Fixed-point \\ weight ($W_0$) \end{tabular} & [0, 1] & 0.5 & 0.244  &N/A & N/A  \\ \hline
\begin{tabular}[c]{@{}c@{}}Scaling functions \\ ($c_+, c_-$)\end{tabular} & (0,1]$\times$(0,1] & \begin{tabular}[c]{@{}c@{}} 0.5, 0.45 \\ ($\frac{c_+}{c_-} = 1.11$)\end{tabular} & \begin{tabular}[c]{@{}c@{}}0.752 , 0.788 \\ ($\frac{c_+}{c_-} = 0.954$) \end{tabular}  & \begin{tabular}[c]{@{}c@{}} 0.5, 0.45 \\ ($\frac{c_+}{c_-} = 1.11$)\end{tabular} & \begin{tabular}[c]{@{}c@{}} 0.894, 0.652 \\ ($\frac{c_+}{c_-} = 1.371$)\end{tabular}\\ \hline
\begin{tabular}[c]{@{}c@{}}Time Constants\\ (ms) ($\tau_+, \tau_-$)\end{tabular} & [10,20]$\times$[20, 40] & 15,30 & 17, 36    & 15,30 & 18,22 \\ \hline \hline
\textbf{Testing Accuracy} & & 91.41 & 90.65  & 88.68 & 89.77\\ \hline
\textbf{Generalization Error} & & 5.79 & 3.17  & 8.29 & 4.61 \\ \hline
\end{tabular}%
}
\end{table}

\begin{figure}
    \centering
    \includegraphics[width = 0.7\columnwidth]{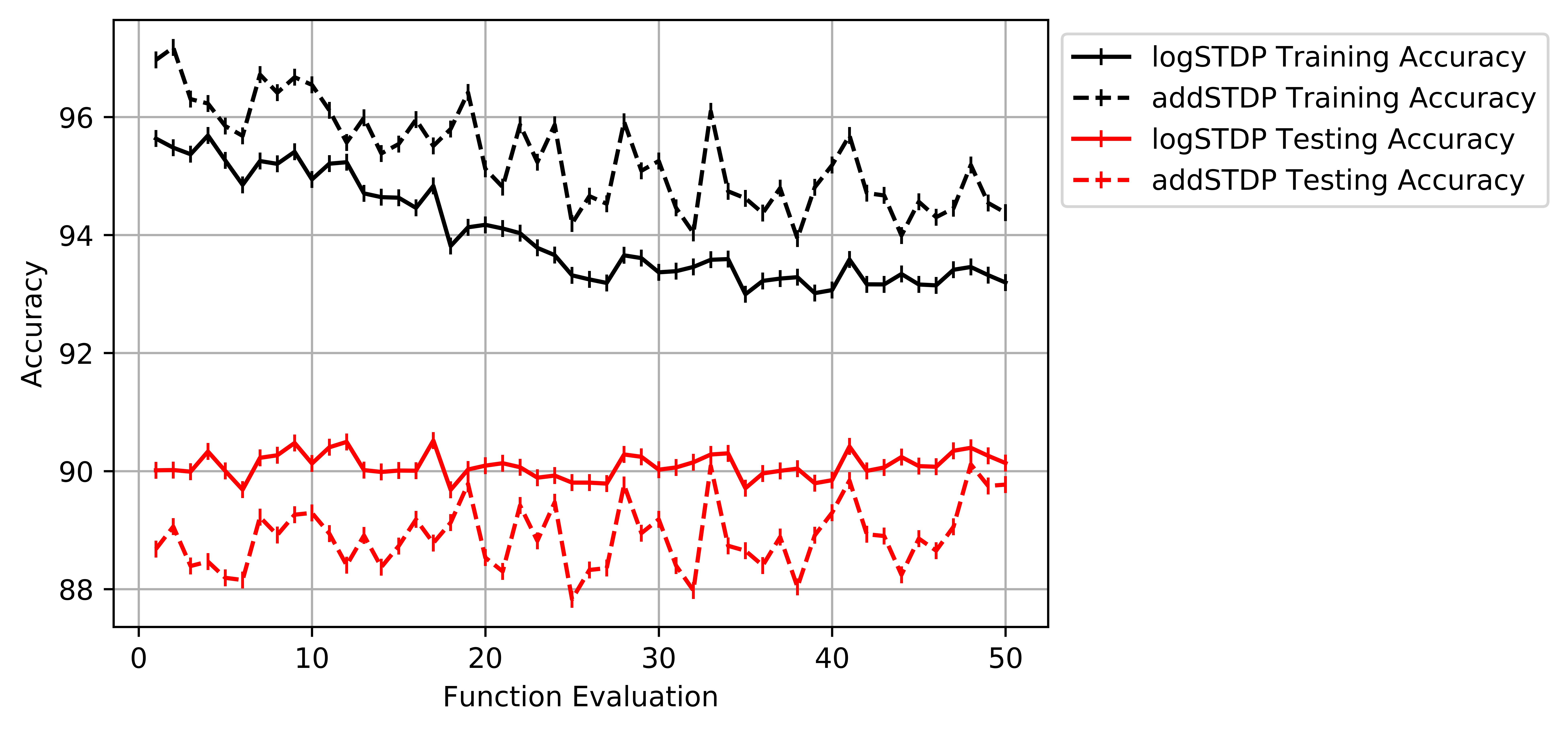}
    \caption{Plot showing comparison between the add-STDP and log-STDP during Bayesian Optimization}
    \label{fig:log_add_bo}
\end{figure}

\textbf{Comparison with Add-STDP:} In order to compare the performance of the log-STDP, we performed a similar analysis using the add-STDP model. The results of the Bayesian Optimization for the add-STDP and the log-STDP are plotted in Figure \ref{fig:log_add_bo}. We see that the log-STDP process outperforms the add-STDP model in terms of both training/testing accuracy and the generalization error thus showing the robustness of the log-STDP process. We see that though the add-STDP has a higher training accuracy, and a comparable test accuracy, its generalization error is higher compared to the log-STDP method.

\section{DISCUSSION}
In this paper, we presented the generalization properties of the spike \hlt{timing}-dependent plasticity (STDP) models. \hlt{A learning process is said to be more generalizable if it can extract features that can be transferred easily to unknown testing sets thus decreasing the performance gap between the training and testing sets. } We provide a theoretical background for the motivation of the work treating the STDP learning process as a stochastic process (an Ornstein-Uhlenbeck process) and modeling it using a stochastic differential equation. We control the hyperparameters of the learning method and empirically study their generalizability properties using the Hausdorff dimension as a measure. \hlt{From Tables \ref{tab:log_scaling},\ref{tab:log_lr} and corresponding Figures \ref{fig:table_plots}, we} observed that the Hausdorff Dimension is a good measure for the estimation of the generalization error of an STDP-trained SNN. We compared the \hlt{generalization error and testing error} for the log-STDP, add-STDP, and mult-STDP models, \hlt{as shown in Tables \ref{tab:log_scaling}, \ref{tab:log_lr}. We} observed that the lognormal weight distribution obtained from the log-STDP learning process leads to a more generalizable STDP-trained SNN \hlt{with a minimal decrease in testing accuracy. 
 In this paper, when we refer to a model as more generalizable, we mean there is a smaller difference between the training and testing performance, i.e., the generalization error. The objective of the paper was to get a model which is more generalizable in the sense that the performance of the network on unknown datasets should not differ much from its performance in the training dataset. It is to be noted that in this paper we are using the generalization error as the metric of generalizability of the network. Generalization error is not a measure of  absolute accuracy, but rather the difference between training and testing datasets. As such, we see that models which have lower generalization error extract  lesser and more important features compared to less generalizable models. However, we see that with this reduced set of features, the model has almost no drop in the testing accuracy, showing the generalizability of the model at comparable accuracy. Thus, we get a model which is more generalizable in the sense that the performance of the network on unknown datasets does not differ much from its performance in the training dataset. As such, these 'more generalizable' models, extract lesser and more important features compared to less generalizable models. However, we see that with this reduced set of features, the model has almost no drop in the testing accuracy, showing the generalizability of the model as we can see from Tables 2, 3. This phenomenon can be explained using the observations of Panda et al. \cite{panda2017asp} on how the ability to forgets boosts the performance of spiking neuron models. The authors showed that the final weight value towards the end of the recovery phase is greater for the frequent input. The prominent weights will essentially encode the features that are common across different classes of old and new inputs as the pre-neurons across those common feature regions in the input image will have frequent firing activity. This eventually helps the network to learn features that are more common with generic representations across different input patterns. This extraction of more generalizable features can be interpreted as a sort of regularization wherein the network tries to generalize over the input rather than overfitting such that the overall accuracy of the network improves. However, due to this regularization, we see that the training performance of the network decreases. However, since the model is more generalizable, the testing performance remains almost constant as seen in Figures 10, 11. 
 } We further observe that the log-STDP models which have a lower Hausdorff dimension and hence have lower generalization error, have a worse trainability i.e., takes a long time to converge during training and also converges to a higher training loss. The observations show that an STDP model can have a trade-off between generalizability and trainability. Finally, we present a Bayesian optimization problem that minimizes the Hausdorff dimension by controlling the hyperparameter of a log-STDP model leading to a more generalizable STDP-trained SNN. 
\par Future work on this topic will consider other models of STDP. In particular, the stochastic STDP rule where the probability of synaptic weight update is proportional to the time difference of the arrival of the pre and post-synaptic spikes has shown improved accuracy over deterministic STDP studied in this paper. The trajectories of such a stochastic STDP model will lead to a Feller process as shown by Kuhn ~\citep{helson2017new}. Hence, in the future, we will perform a similar Hausdorff dimension-based analysis for generalization for the stochastic STDP model. Moreover, in this work, we have only considered the hyperparameters of the STDP model to improve the generalizability of the SNN. An important extension is to consider the properties of the neuron dynamics, which also controls the generation of the spikes and hence, weight distribution. The choice of the network architecture will also play an important role in the weight distribution of the SNN. Therefore, a more comprehensive optimization process that couples hyperparameters of the STDP dynamics, neuron dynamics, and network architecture \hlt{like convolutional SNN \citep{kheradpisheh2018stdp} and heterogeneous SNN \citep{she2021heterogeneous}} will be interesting future work.

\section*{Funding}
This material is based on work sponsored by the Army Research Office and was accomplished under Grant Number W911NF-19-1-0447. The views and conclusions contained in this document are those of the authors and should not be interpreted as representing the official policies, either expressed or implied, of the Army Research Office or the U.S. Government.





\bibliographystyle{frontiersinSCNS_ENG_HUMS} 
\bibliography{test}

\end{document}